\documentclass[letterpaper, 10 pt, conference]{ieeeconf}  % Comment this line out if you need a4paper

\IEEEoverridecommandlockouts                              % This command is only needed if 
\usepackage{comment}
\usepackage{amsmath}
\usepackage{amssymb}
\usepackage[dvipsnames]{xcolor}
\usepackage{xspace}
\usepackage{algorithm}
\usepackage[noend]{algpseudocode}
\usepackage{graphicx}
\usepackage{subcaption} 
\usepackage{svg}
\usepackage{gensymb}
\usepackage{ragged2e}
\usepackage{cite}
\usepackage{soul}
\usepackage[pdfa,colorlinks,bookmarksopen,bookmarksnumbered,allcolors=blue]{hyperref}
\usepackage[normalem]{ulem} % only for \sout in used by Zach

\newcommand{\pname}{TIC\,\,}
\newcommand{\pnamens}{TIC}

\title{\LARGE \bf
Beyond Fixed Goal Delivery: Online POMDP Planning for Target Interception in Crowds
}

\author{
Himanshu Gupta${^{1}}$, Kelvin Aladum${^{1}}$, Nisar Ahmed${^{1}}$,
Bradley Hayes${^{2}}$ and Zachary Sunberg${^{1}}$
\thanks{This work was supported by NSF Awards \#2133141 and \#2137269
and by members of the Center for Autonomous Air Mobility and
Sensing IUCRC.}
\thanks{
$^{1}$Authors are with the Dept. of Aerospace Engineering Sciences, and
$^{2}$Author is with the Dept. of Computer Science,
University of Colorado Boulder, USA.
{\tt Emails:\{firstname.lastname\}@colorado.edu}
}
\thanks{\copyright~2026 IEEE. Personal use of this material is permitted.
Permission from IEEE must be obtained for all other uses, in any current or
future media, including reprinting/republishing this material for advertising
or promotional purposes, creating new collective works, for resale or
redistribution to servers or lists, or reuse of any copyrighted component of
this work in other works.}
}

\begin{document}

\maketitle
\thispagestyle{empty}
\pagestyle{empty}

%%%%%%%%%%%%%%%%%%%%%%%%%%%%%%%%%%%%%%%%%%%%%%%%%%%%%%%%%%%%%%%%%%%%%%%%%%%%%%%%
\begin{abstract}

Target interception in crowded environments requires reaching a moving objective while navigating among multiple uncertain human agents.  
Since human navigation intent is not directly observable, the robot must reason over multiple possible future interaction outcomes.
We formulate interception in crowds as a partially observable Markov decision process and solve it online using tree search under a fixed computational budget.
In this setting, the action-space structure directly shapes the search tree and how computational effort is allocated.
We perform a controlled comparison between a sequential path–speed planner, which first plans a spatial path and then modulates speed along it, and a unified planner that jointly branches over steering and speed within tree search.
Across simulations with up to 200 humans, both approaches perform similarly at low crowd density but diverge sharply as density increases.
At the highest crowd density, the sequential planner has a safe-interception rate 31 percentage points lower and requires 44\% more time than the unified steering--speed planner, revealing a structural limitation of spatial restriction.
\textcolor{blue}{\href{https://tic-planning.github.io/}{Project Webpage: \tt tic-planning.github.io}}
\end{abstract}
%%%%%%%%%%%%%%%%%%%%%%%%%%%%%%%%%%%%%%%%%%%%%%%%%%%%%%%%%%%%%%%%%%%%%%%%%%%%%%%%
\section{Introduction}

\begin{figure}[t]
    \centering
    \includegraphics[width=0.98 
    \linewidth]{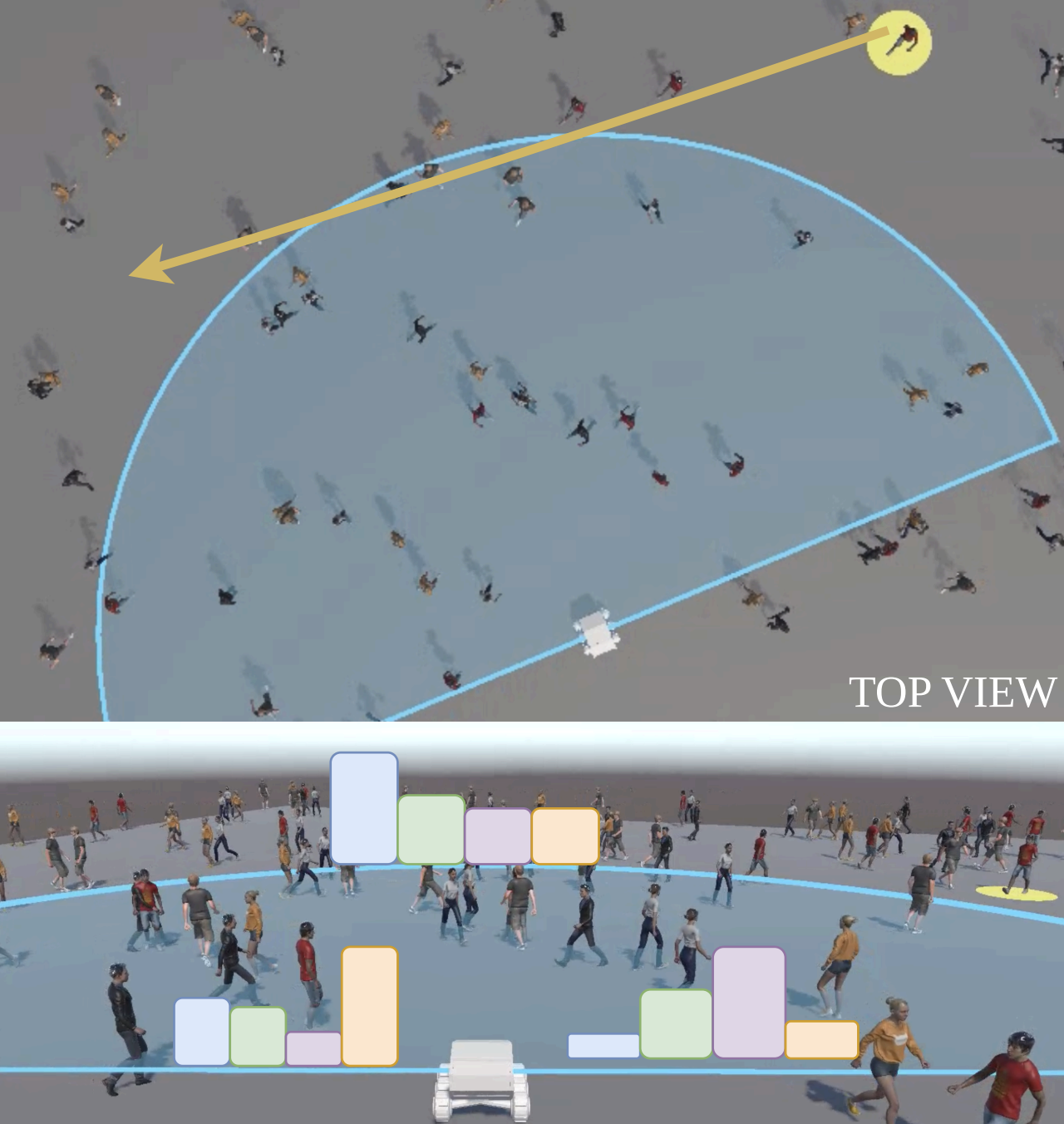}
    \caption{
    \small
    Target Interception in Crowds (TIC). A robot (white) must intercept a moving target (yellow) while navigating among many intention-driven humans. The robot observes nearby humans within a limited sensing region (blue region) and maintains a belief over their intent hypotheses (colored bars), requiring planning under partial observability. The simulated environment is shown from a top-down view (top) and a perspective view (bottom).
}
    \label{fig_intro_image}
\end{figure}

Autonomous robots increasingly operate in human-populated environments such as hospitals and airports.
Consider a delivery robot attempting to reach a doctor moving between patient rooms, or navigating a dense airport crowd to hand an order directly to a customer.
Traditional delivery systems assume a fixed drop-off location and require the recipient to wait there. 
In contrast, a more efficient strategy is for the robot to actively navigate the environment and intercept the recipient to complete the handover.
We refer to this problem as Target Interception in Crowds (\pnamens) (Fig.~\ref{fig_intro_image}).

Prior work on reaching a moving target has largely considered static environments~\cite{zheng2021time,gupta2022shortest} or settings with known dynamic obstacles~\cite{Qu2022}.
Navigating toward a moving target in crowds introduces an additional layer of complexity, since surrounding humans behave as uncertain dynamic agents.
A common modeling approach represents each human as an intention-driven agent pursuing a latent goal that is not directly observable to the robot~\cite{luo2018porca}.
This enables forward simulation of human motion, but introduces the additional challenge of inferring each human’s intent from observations.
The robot must therefore pursue a moving target while reasoning over human intent uncertainty, subject to its own motion constraints.
This decision-making problem requires planning under partial observability, which we model as a Partially Observable Markov Decision Process (POMDP).

While this POMDP formulation captures uncertainty arising from hidden human intent, solving it exactly is computationally intractable~\cite{pomdp_complexity_papadimitriou_1987}.
In practice, robotic systems rely on approximate methods for solving POMDPs, including learning-based methods~\cite{letsdrive,lee2020magic}, model predictive control (MPC)~\cite{hu2022sharp,mustafa2024racp}, and sampling-based tree search~\cite{lauri2022partially}, each with distinct tradeoffs discussed in Section~\ref{sec_related_work}.
In this work, we adopt an online tree search-based POMDP planning approach.
For tree search methods, the action space directly determines the branching factor and the set of reachable future states.
Under limited computational budgets, the planner evaluates only a sparse subset of this space, so the structure of the action space strongly influences which strategies can be discovered.
This raises an important question: how does action-space structure influence the effectiveness of online tree-search planning for interception tasks in dense crowds?

For car-like robots navigating in dynamic environments, a common planning strategy is to first compute a feasible path to the objective and then optimize speed along it.
This path-first decomposition appears both in navigation among humans~\cite{bai2015intention,luo2018porca,importance_sampling_despot} and in interception literature~\cite{Qu2022}.
We refer to this formulation as Limited Space Planning (LSP).
By restricting tree search to speed control along a fixed path, LSP reduces the branching factor and permits deeper exploration under a fixed computational budget.
Prior work has shown that for navigation to a static goal among humans, allowing tree search to control both steering and speed leads to safer and more efficient behavior than LSP~\cite{letsdrive,gupta_ped_navigation_holonomic,gupta2024efficient}.
This formulation, referred to as Extended Space Planning (ESP), expands the reachable state space and enables the planner to modify the robot’s spatial trajectory during search.

However, dynamic interception introduces a new structural tension. 
Unlike fixed-goal navigation, interception requires the robot to meet the target at a compatible time along its trajectory.
Because crowd-induced delays dictate where and how interception can occur, timing and spatial decisions become tightly and inextricably coupled.
Deeper search along a predicted interception path, as in LSP, may improve long-horizon timing reasoning.
In contrast, increased spatial flexibility, as in ESP, may better preserve maneuvering options when delays require significant changes in spatial trajectory.
It is therefore unclear whether deeper search or greater spatial expressiveness yields better performance as crowd density increases.

\textbf{\emph{Contributions:}} 
In this work, we study the role of action-space structure in online tree search for dynamic target interception in crowded environments.
We perform a controlled comparison between Limited and Extended Space Planning (LSP vs. ESP) under identical sensing, belief estimation, reward structure, and computational budgets, thereby isolating the impact of spatial restriction versus flexibility.
Through extensive simulations across increasing crowd densities, we demonstrate that path-restricted planning degrades significantly under dense multi-agent interaction.
While LSP and ESP perform comparably at low crowd density, the performance gap widens as density increases, with LSP degrading faster in terms of safe interception rate and motion efficiency.
These results reveal a structural limitation of spatially constrained search and establish the importance of action-space expressiveness when intercepting moving targets in dense multi-agent settings.

While our primary study assumes a known target state, we additionally evaluate a partially observable target setting in which the target state is estimated using a particle filter and the planner tries to intercept the most likely target estimate.
This secondary study examines whether the structural advantages of ESP persist under estimation error.
%%%%%%%%%%%%%%%%%%%%%%%%%%%%%%%%%%%%%%%%%%%%%%%%%%%%%%%%%%%%%%%%%%%%%%%%%%%%%%%%
\section{Related Work} \label{sec_related_work}

Target interception in crowded environments combines two challenges: intercepting a moving objective and navigating safely among many uncertain human agents. Existing work typically addresses these components in isolation.

\emph{Interception with simplified environments:}
A large body of interception work assumes static environments.
Related work derives candidate interception solutions for curvature-bounded vehicles under simplified target motion models, including time or length optimal Dubins interception variants~\cite{zheng2021time,gupta2022shortest}, and aerial capture systems that execute selected intercept maneuvers or proportional-navigation guidance~\cite{pliska2024towards}.
Qu et al.~\cite{Qu2022} follow a predict-then-intercept pipeline for interception among known, deterministic dynamic obstacles. 
They predict the target trajectory, select an intercept point, plan a feasible path to the point, and use a space--time graph for speed control along that path.
These approaches perform well when target motion can be predicted reliably, and environmental constraints remain relatively stable, but they do not address interception in dense, intention-driven crowd interaction.

\emph{Crowd navigation under intent uncertainty:}
Bai et al.~\cite{bai2015intention} introduced online POMDP planning, applied using an LSP formulation, for the purpose of reasoning over latent human intent hypotheses while navigating in crowds.
Cai et al.~\cite{letsdrive} extended this framework to ESP, demonstrating improved performance through learned value approximations.
While learning-based approximations can improve computational efficiency, their accuracy depends on training distribution and may degrade in rare or safety-critical scenarios~\cite{xu2023benchmarking,raj2024rethinking}.
In contrast, model-based online planning evaluates actions directly through forward simulation, allowing adaptation to unexpected behaviors within the fidelity of the predictive model~\cite{jin2025hi}.
Other online approaches~\cite{hu2022sharp,peters2024contingency,mustafa2024racp} that consider multiple human-intent hypotheses also optimize for heading and speed changes jointly.  
Using an MPC formulation sometimes referred to as open-loop feedback control (OLFC)~\cite{bertsekas2005dynamic}, each time a feedback observation is received, these methods compute a single optimized open-loop trajectory.
By contrast, online tree-search formulations plan closed-loop policies by branching over possible human responses, enabling belief-conditioned reasoning over multiple futures.
Gupta et al.~\cite{gupta_ped_navigation_holonomic} demonstrate that online tree search with ESP achieves strong performance using effective rollout policies without relying on learned value functions.
These works navigate to a known fixed goal.
This assumption breaks in target interception, where the objective is dynamic, and the interception point evolves during execution.

\emph{Active tracking and pursuit--evasion:}
Active tracking methods plan actions to reduce posterior uncertainty over the target or improve observability~\cite{schlotfeldt2019maximum,yang2023policy}.
Pursuit--evasion and visibility-based planning study capture or detection under adversarial or occlusion constraints, often with few agents and static obstacles~\cite{olsen2021visibility,olsen2022robust}.
While these threads are related, their primary objectives and scaling regimes differ from interception in dense crowds under intent uncertainty.

\emph{Gap addressed in this work.}
Prior interception pipelines typically ignore crowd intent uncertainty, while crowd-navigation methods that study action-space structure in tree search assume fixed goals.
Crucially, prior work has not examined how the structural assumptions of decoupled path-and-speed planning break down in dynamic target interception as crowd density scales.
This motivates a controlled study of LSP and ESP within an online tree-search planning framework for dynamic target interception in crowds.
%%%%%%%%%%%%%%%%%%%%%%%%%%%%%%%%%%%%%%%%%%%%%%%%%%%%%%%%%%%%%%%%%%%%%%%%%%%%%%%%
% \input{sections/problem_statement}
%%%%%%%%%%%%%%%%%%%%%%%%%%%%%%%%%%%%%%%%%%%%%%%%%%%%%%%%%%%%%%%%%%%%%%%%%%%%%%%%
\section{TIC POMDP Problem Formulation} \label{sec_pomdp_formulation}

\subsection{POMDP Preliminaries}
A Partially Observable Markov Decision Process (POMDP) provides a principled framework for sequential decision-making under uncertainty.
A POMDP is defined by the tuple $(S, A, Z, T, O, R, \gamma)$.
Here, $S$ is the state space, $A$ is the action space, $Z$ is the observation space, $T(s,a,s')$ is the transition model, $O(s',a,z)$ is the observation model, $R(s,a)$ is the reward function, and $\gamma \in (0,1)$ is the discount factor.
The true state, $s \in S$, is not directly observable.
The robot therefore maintains a belief state $b$, which is a probability distribution over $S$, with $b(s_c)$ indicating the likelihood of $s_c$ being the true state.
Let $b_{t}$ denote the belief state at time $t$. 
If the system takes an action $a_t$ and gets an observation $z_{t+1}$ at the next time step $t+1$, then using Bayes' rule, we get the new belief state $b_{t+1}$ as:
\begin{align} \label{belief_update}
b_{t+1}(s') \propto O(s', a_t, z_{t+1}) \sum_{s\in S} T(s, a_t, s') b_{t}(s)
\end{align}

The objective of any POMDP solution technique is to find a policy $\pi^*$ that maximizes the expected discounted cumulative reward, i.e., $\pi^* = \arg\max_{\pi} V^{\pi}(b)$ where

\begin{align}
    V^{\pi}(b) &= \mathbb{E} \left[ \sum_{t=0}^{\infty} \gamma^t R(s_t, \pi(b_t)) \mid b_0 = b \right]
\end{align}

\subsection{\pname as a POMDP} \label{sec_problem_formulation}

\subsubsection{State Space} \label{sec_pomdp_state}
The state consists of three components:
(i) the robot state $s_R$,
(ii) the target state $s_T$, and
(iii) the states of $N$ nearby humans $\{s_H^i\}_{i=1}^{N}$.
The full state is $s = (s_R, s_T, s_H^1, \ldots, s_H^N)$.
The robot and target state encode their position and motion variables. 
Each human state $s_H^i$ includes position, motion variables, and the intent variable $g_i$.
The human intent variables $g_i$ are not directly observable, inducing partial observability in the system.

\subsubsection{Action Space} \label{sec_pomdp_action}
The action space $A$ consists of control inputs admissible under the robot dynamics.
At each decision step, the planner selects an action $a \in A$, which is executed for duration $\Delta t$.
Section~\ref{sec:action_space_formulations} describes the action-space formulations used in this work.
Additionally, the action space includes a sudden braking (SB) action that enables rapid deceleration in response to imminent collision. 

\subsubsection{Observation Space} \label{sec_pomdp_observation}
At each time step, the robot receives an observation $z \in Z$.
The robot directly observes its own state, the target state, and the positions of nearby humans without noise.
While realistic sensors do exhibit noise, this modeling choice isolates the impact of latent intent uncertainty from localization errors, consistent with established crowd navigation literature ~\cite{bai2015intention,luo2018porca,importance_sampling_despot,gupta_ped_navigation_holonomic}.
This modeling choice simplifies belief updates but is not required by the POMDP framework.

\subsubsection{Reward Function} \label{sec_pomdp_reward}

The reward function balances interception success, safety, and efficiency.
A positive terminal reward $R_{\mathrm{I}}$ is given when the robot reaches the target within a capture radius $d_c$. 
Collisions with the environment boundary incur a negative reward $R_{\mathrm{obs}}$. 
For human interactions, we adopt the not-at-fault safety formulation of~\cite{vaskov2019not}. 
If the robot violates a minimum safety distance $d_s$ to any human while moving, it incurs a penalty $R_{\mathrm{h}}$, whereas no penalty is applied when the robot is stationary.
To discourage unnecessary emergency braking, selecting the \textsc{SB} action results in a penalty $R_{\mathrm{SB}}$.
Finally, a constant per-step penalty $R_{\mathrm{step}}$ promotes faster interception. The total reward is the sum of these terms, with terminal rewards applied upon interception or collision.

\subsubsection{Generative Model} \label{sec_pomdp_genfunc}
We assume access to a simulator-based generative model $G(s,a)$ that returns a sampled next state $s'$, observation $z$, and reward $r$.
Given state $s=(s_R,s_T,\{s_H^i\})$ and action $a$, the system evolves over one decision interval $\Delta t$ as follows:
$
% \[
s_R' = f_R(s_R,a), \,\,
s_T' = f_T(s_T), \,\,
s_H^{i\prime} = f_H(s_H^i,g_i) + w_H^i
% \]
$.
$f_R$ and $f_T$ denote the robot and target motion models, and $f_H$ is the human motion model parameterized by intent. 
Robot and target dynamics are deterministic during planning, while human motion is stochastic with bounded process noise $w_H^i$.
An observation $z$ is generated according to $z = h(s')$, where $h$ is defined in Section~\ref{sec_pomdp_observation}.
The reward $r$ is computed from $(s,a,s')$ using the reward function defined in Sec.~\ref{sec_pomdp_reward}.
This generative formulation enables sampling-based online planning without explicit representations of transition or observation probabilities, as is standard in online POMDP solvers such as ARDESPOT~\cite{despot} and POMCPOW~\cite{pomcpow}.

\subsection{Belief Representation} \label{sec_belief_representation}

The robot maintains a belief over the intent variables of nearby humans.
We assume the robot state, target state and human positions are directly observable without noise, and therefore do not require belief representation.

\emph{Human Intent Belief:}
Each human $i$ has a discrete intent variable $g_i \in \mathcal{G}$, referred to as its goal, where $\mathcal{G}$ is a finite set of candidate goal locations.
For each human, the robot maintains a categorical distribution $b_i(g)$ over $g \in \mathcal{G}$.
When a human is first observed, the belief $b_i(g)$ is initialized uniformly over all the goal locations.
For previously observed humans, the belief is updated using Bayes' rule.
The likelihood term is derived from the intent-conditioned human motion model $f_H(\cdot, g)$, which predicts motion under goal $g$.
Intents whose predicted motion better explains the observed displacement receive higher posterior probability.

%%%%%%%%%%%%%%%%%%%%%%%%%%%%%%%%%%%%%%%%%%%%%%%%%%%%%%%%%%%%%%%%%%%%%%%%%%%%%%%%
\section{Solution Approach} \label{sec_solution_approach}

\paragraph*{System Overview}

The system operates in an online replanning loop with a control period $\Delta t$.
At time $t$, the robot knows its current state, the target state and is executing a previously selected control over the interval $[t,\, t+\Delta t)$.
It observes all humans detected within its sensing range and maintains beliefs over their latent intentions.
For planning, only the $N$ nearest humans are incorporated into tree search.
This focuses computational effort on the most safety-critical interactions while limiting search complexity.
Using the current belief at time $t$, the planner performs online tree search for a fixed budget of $\Delta t$ seconds to select the control to be executed over the next interval $[t+\Delta t,\, t+2\Delta t)$.
Thus, planning at time $t$ overlaps with execution over $[t,\, t+\Delta t)$, enabling continuous operation without interrupting control.
The simulated search horizon is sufficiently long for rollouts to reach an interception event whenever reasonably achievable.
At time $t+\Delta t$, the newly selected control is applied, new observations are incorporated to update the beliefs, and the process repeats.
This loop continues until the robot reaches the target within a predefined capture radius.

\subsection{Online Tree Search}

At each planning cycle, the robot uses sampling-based online tree search to approximately solve the \pname POMDP.
Given the current belief, the planner samples possible future realizations of human behavior and propagates the target deterministically under its motion model.
The search incrementally builds a tree of simulated action sequences by invoking the generative function described in Section~\ref{sec_pomdp_genfunc}.
Tree search operates over a finite discretization of the robot’s action space.
The system evolves in a continuous state space with stochastic human dynamics, and tree search operates on sampled uncertainty realizations to approximate future evolution.
Each tree node corresponds to a simulated future state of the system under these sampled realizations.
Leaf nodes are evaluated using rollout simulations to estimate the long-horizon value beyond the explored portion of the tree.
Under a finite computational budget, the resulting search depth is limited, making rollout-based leaf evaluation essential for guiding tree expansion.
After the computational budget is exhausted, the action at the root with the highest estimated value is executed.

\begin{figure}[t!]
    \centering
    \begin{subfigure}[b]{0.4\columnwidth}
        \centering
        \includegraphics[width=0.99\columnwidth]{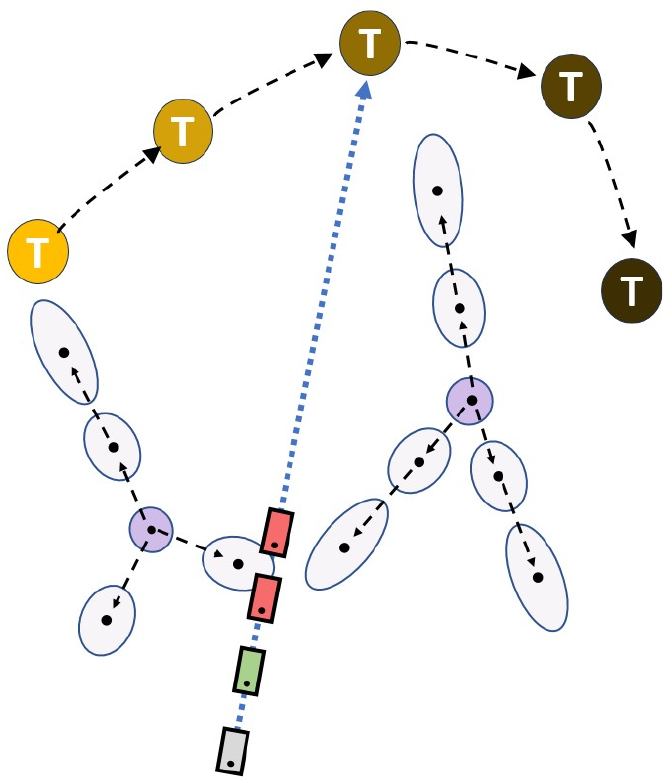}
        % \caption{Limited Space Planner}
        \label{fig_lsp_planning}
    \end{subfigure}%
    ~~~~~
    \begin{subfigure}[b]{0.4\columnwidth}
        \centering
        \includegraphics[width=0.99\columnwidth]{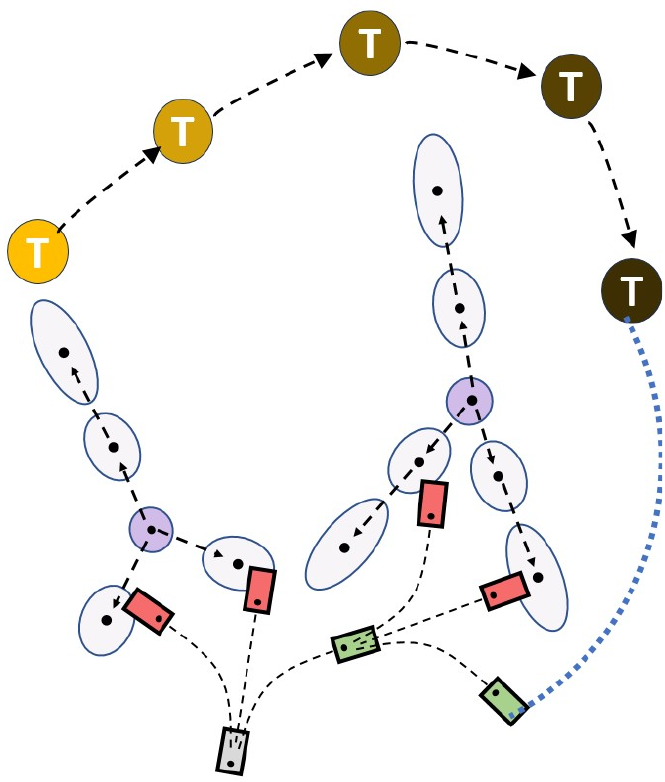}
        % \caption{Extended Space Plan}
        \label{fig_esp_planning}
    \end{subfigure}%
    \caption{
    \small
    Illustration of the search trees generated by the Limited-Space Planning (LSP) and Extended-Space Planning (ESP) formulations.
    LSP (left) finds a path (dotted blue line) to the predicted target position (yellow) and performs tree search along that path.
    The search encounters a robot collision (red) with a human (purple), so the planner chooses to slow down.
    In contrast, ESP (right) explores multiple spatial trajectories, finds collision-free states, evaluates them using rollouts (dotted blue line), and selects a right-turn maneuver to move around the humans.
    }
    \label{fig_planning_trees}
\end{figure}

\subsection{Action-Space Formulations}
\label{sec:action_space_formulations}

The structure of the action space used by tree search determines which future trajectories can be explored under a fixed computational budget.
We consider a car-like robot whose motion is controlled by forward velocity and steering.
We study two formulations that differ in how these controls are represented in the planner's action space (Fig.~\ref{fig_planning_trees}).

\subsubsection{Limited Space Planning (LSP)}

Limited Space Planning~\cite{bai2015intention} adopts a decoupled architecture.
A path-planning algorithm first computes a dynamically feasible path toward a target hypothesis, thereby determining the robot’s heading evolution.
Online POMDP planning performs tree search to reason about uncertainty in human intent and motion while selecting how fast or slow the robot should proceed along the path.
Heading adjustments are not optimized within tree search but are determined by the precomputed path.

\subsubsection{Extended Space Planning (ESP)}

Extended Space Planning~\cite{gupta_ped_navigation_holonomic} eliminates the reference-path constraint used in LSP.
Online POMDP planning directly selects dynamically feasible controls that determine both heading and forward motion, allowing spatial decisions to adapt during search rather than being fixed in advance.
Heading and speed changes are therefore optimized jointly within tree search under uncertainty.

\subsection{Rollout Policy} \label{sec_rollout_policy}

The reward structure of the \pname problem is sparse, with positive reward obtained only upon successful interception.
Under a finite computational budget, tree search may not expand deeply enough to discover interception events.
Effective rollout policies are therefore essential for providing informative value estimates at leaf nodes, which in turn guide tree expansion toward high-value regions of the search space~\cite{gupta_ped_navigation_holonomic,gupta2024efficient,swiechowski2023monte}.
We design rollout policies that promote interception-oriented behavior whenever feasible.
Rollouts terminate upon interception, collision, or when the predefined maximum depth is reached.

\subsubsection{LSP Rollout} \label{sec_lsp_rollout}

Under LSP, tree search is constrained to a reference path.
At a given tree node, the remaining portion of this path determines the robot’s heading evolution during rollout.
Following prior work, we employ a conservative reactive controller to modulate speed along the path~\cite{bai2015intention,luo2018porca,importance_sampling_despot}.
Each tree node contains a simulated system state, including the robot and the $N$ nearby humans.
If no humans are within distance $D_{\mathrm{far}}$ of the robot, the robot increases its speed.
If any human is within distance $D_{\mathrm{near}}$, the robot decreases its speed.
Otherwise, the robot does not change its speed.
This rule promotes cautious behavior in dense regions while allowing faster progress in open space.

\subsubsection{ESP Rollout} \label{sec_esp_rollout}

Under ESP, no reference path is available, so changes in heading and forward motion must be determined during rollout.
The speed update is first determined using the same reactive controller as in LSP (Section~\ref{sec_lsp_rollout}).
Each tree node contains a simulated system state, including the robot and target state.
At every rollout step, the target state is propagated forward for $\Delta t$ using the target dynamics model.
For the robot, each steering command in the action set is combined with the selected speed update and simulated forward for $\Delta t$.
The action that brings the robot closest to the propagated target state is selected; this ensures rollout actively reduces distance to the target at each step, whether or not interception occurs during tree expansion.

To ensure a fair comparison, both LSP and ESP rollouts utilize identical reactive speed modulation based on human proximity. The divergence in their spatial behavior is a direct, unavoidable consequence of their respective action-space formulations rather than an injected bias.

%%%%%%%%%%%%%%%%%%%%%%%%%%%%%%%%%%%%%%%%%%%%%%%%%%%%%%%%%%%%%%%%%%%%%%%%%%%%%%%%
\section{Experiments} \label{sec_experiments}

We evaluate the planners in a dynamic target interception setting across increasing crowd densities.

\subsection{Simulation Environment and System Models}
We compare the different action-space formulations for \pname in simulation. 
The simulation environment is a $50\,\mathrm{m} \times 50\,\mathrm{m}$ workspace without static obstacles.
State variables for the robot, target, and humans are expressed in a common global Cartesian frame.
We use variables $x,y$ to denote position, $\theta$ to denote heading and $v$ to denote speed.

\emph{Robot Model:}
The car-like robot follows a kinematic bicycle model.
Its state is $s_R = (x_R, y_R, \theta_R, v_R)$.
The control inputs consist of steering angle $\phi_R$ and a speed update $\delta_v$, where the commanded speed at the next decision step is given by
$v_R \leftarrow \mathrm{clip}(v_R + \delta_v,\, 0,\, v_{m})$.
The continuous-time dynamics evolve as
$
\dot{x}_R\!=\!v_R \cos\theta_R, 
% \quad
\;
\dot{y}_R\!=\!v_R \sin\theta_R,
% \quad
\;
\dot{\theta}_R\!=\!\frac{v_R}{L}\tan\phi_R,
$
where $L\!=\!0.75\,\mathrm{m}$ is the wheelbase.
The maximum allowed speed is $v_{m}\!=\!2\,\mathrm{m/s}$ and the maximum allowed steering angle is $\phi_m\!=\!0.475$\,$\mathrm{rad}$.

\emph{Target Model:}
The target state is $s_T\!=\!(x_T, y_T, \theta_T, v_T)$.
The target evolves according to a constant speed and turn-rate model given by
% \[
$
\dot{x}_T\!=\!v_T \cos\theta_T, \;
% \quad
\dot{y}_T\!=\!v_T \sin\theta_T, \;
% \quad
\dot{\theta}_T\!=\!\omega,
$
% \]
where $\omega\!=\!0.05\,\mathrm{rad/s}$. 
When approaching workspace boundaries, the target heading is smoothly redirected to remain within bounds.

\emph{Human Model:}
Each human $i$ has state 
$s_H^i\!=\!
(x_H^i, y_H^i, \theta_H^i, v_H^i, g_i)$, 
where $g_i$ is the goal location representing the human’s intent.
Every human moves at a constant speed $v_H^i$ and follows a goal-directed model in which the heading $\theta_H^i$ at each step points toward the goal location $g_i$. 
The dynamics are
$
% \[
\dot{x}_H^i\!=\!v_H^i \cos\theta_H^i, 
% \quad
\;
\dot{y}_H^i\!=\!v_H^i \sin\theta_H^i.
% \]
$

\emph{Observation Model:}
At each time step, the robot observes its own state, the target state, and the positions of all humans within a $15\,\mathrm{m}$ sensing radius and 360$^{\circ}$ field of view.
These observations are assumed to be perfectly measured, consistent with prior navigation work~\cite{bai2015intention,luo2018porca,importance_sampling_despot,gupta_ped_navigation_holonomic}.

\emph{Belief Configuration:}
For human intent, a categorical belief is maintained over the goal hypotheses.
Four corners of the workspace serve as candidate goal locations for humans.

Robot and target dynamics are deterministic during simulation to isolate the effects of crowd-induced uncertainty on interception performance. 
Human motion is stochastic, with bounded additive displacement noise $w_H^i \sim \mathcal{U}(-0.2,\, 0.2)\,\mathrm{m}$ applied along the direction of motion at each time step to capture behavioral variability.
The proposed POMDP framework readily accommodates additional sources of uncertainty (e.g., robot actuation noise or stochastic target motion), as well as alternative robot models (e.g., second-order car), alternative target dynamics (e.g., intent-driven models),  and alternative human motion models (e.g., Social Force~\cite{helbing1995social} or PORCA~\cite{luo2018porca}), without modification.

\subsection{Online POMDP Planner}

We use the ARDESPOT~\cite{despot} algorithm as implemented in \texttt{POMDPs.jl}~\cite{egorov2017pomdps} to solve the POMDP online.
Planning operates in a receding-horizon loop with decision period $\Delta t = 0.5\,\mathrm{s}$.
Since ARDESPOT struggles with large
or continuous action space problems, tree search is performed over a discretized action space.
Developing online POMDP solvers that can better solve continuous action
space problems is an active area of research~\cite{vomcpow}.

\emph{Scenario Initialization:}
At each planning step, ARDESPOT samples $K\!\!=\!\!50$ scenarios from the current belief. 
For each of the $N\!=\!6$ nearest humans, intent is sampled from its categorical belief.
The target state in all scenarios is initialized to the observed target state at the current time step.

\begin{figure*}[t!]
    \centering
    \begin{subfigure}[t]{0.25\textwidth}
        \centering
        \includegraphics[width=\linewidth]{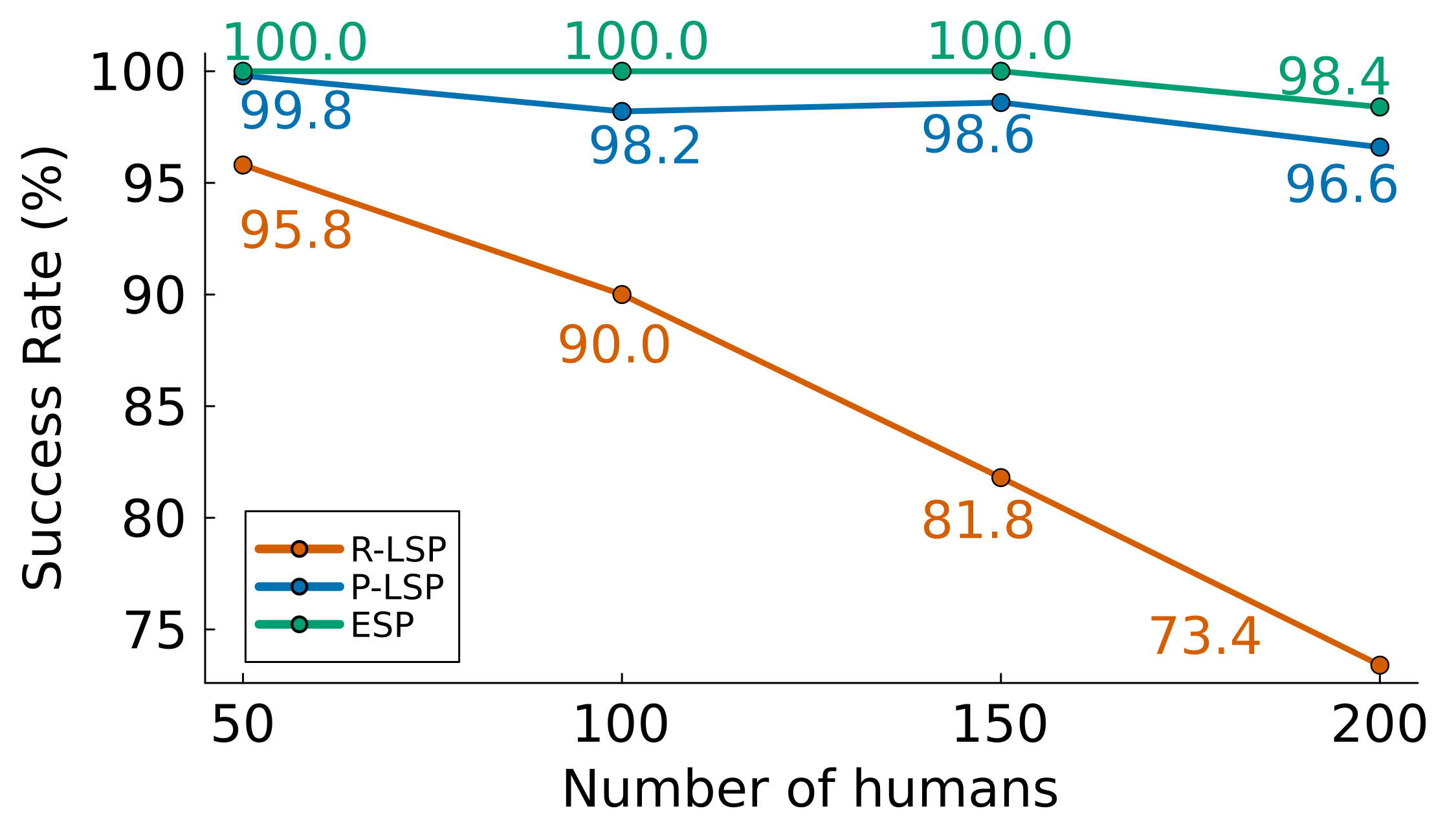}
        \caption{\small Success Rate}
        \label{fig_results_primary_success_rate}
    \end{subfigure}\hfill
    \begin{subfigure}[t]{0.25\textwidth}
        \centering
        \includegraphics[width=\linewidth]{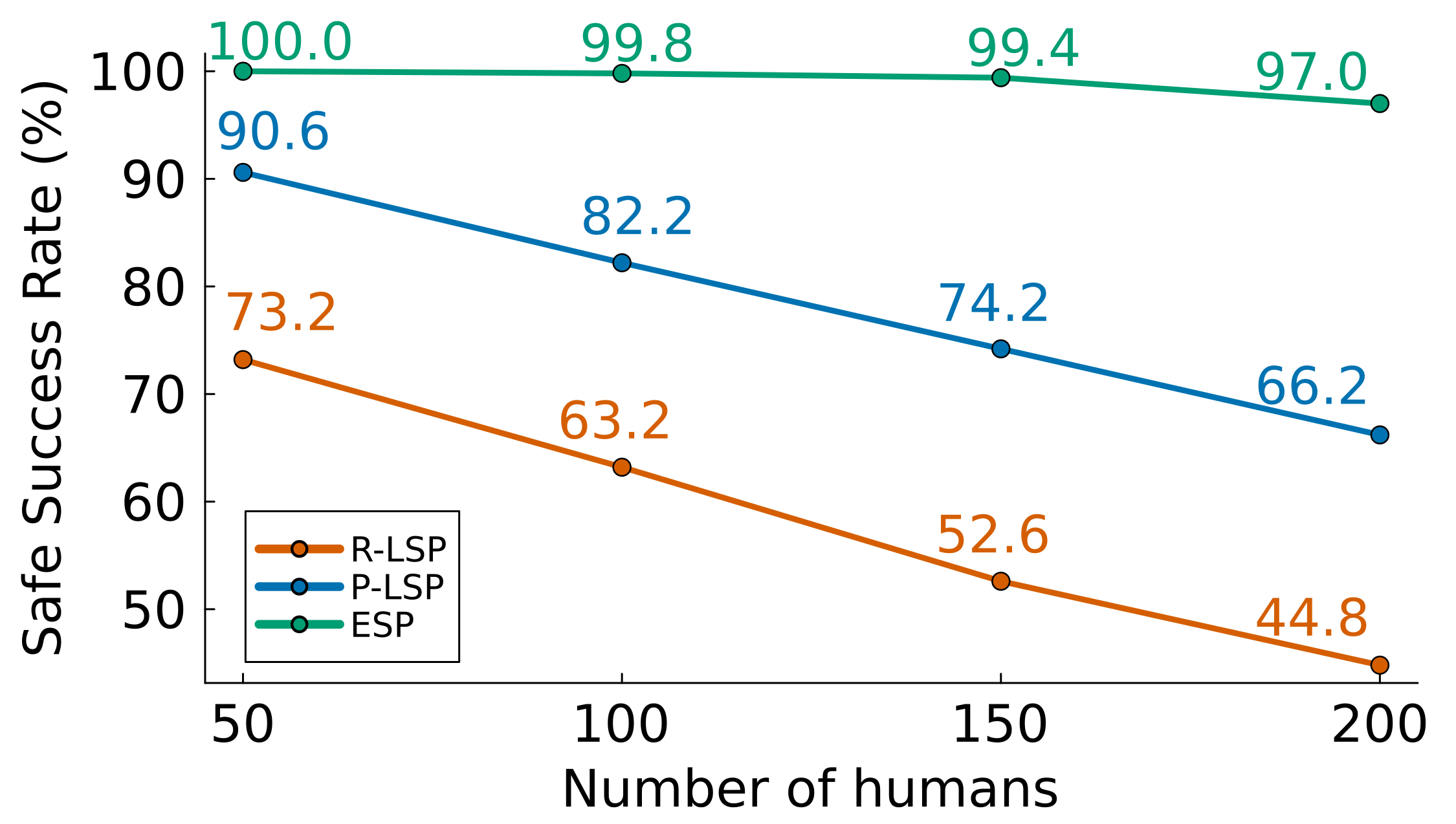}
        \caption{\small Safe Success Rate}
        \label{fig_results_primary_safe_success_rate}
    \end{subfigure}\hfill
    \begin{subfigure}[t]{0.25\textwidth}
        \centering
        \includegraphics[width=\linewidth]{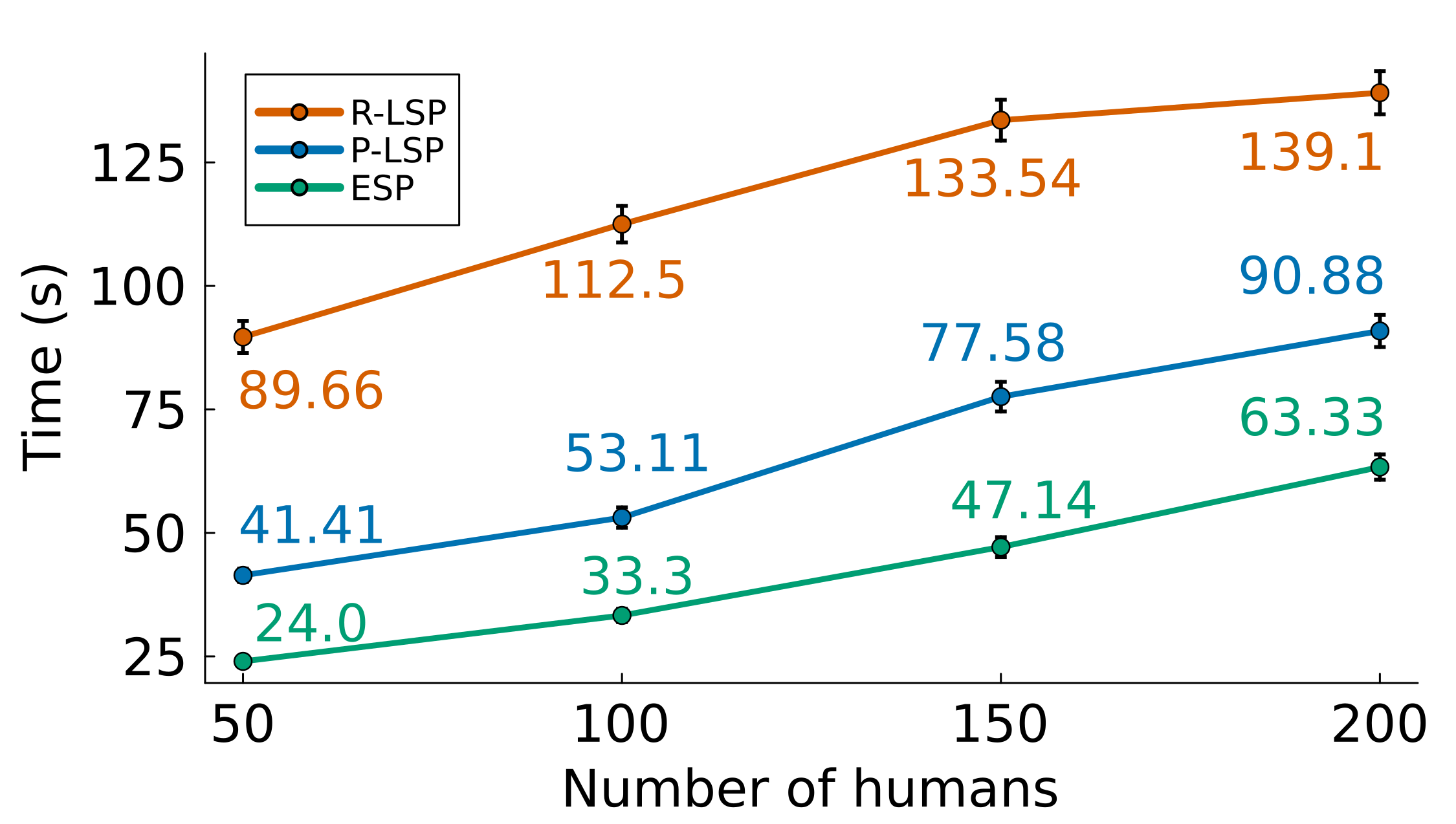}
        \caption{\small Time to Interception}
        \label{fig_results_primary_interception_time}
    \end{subfigure}
    \caption{\small Primary study (fully observable target). Interception success rate, safe-success rate (zero proximity violations), and time-to-interception as functions of crowd density.
    }
    \label{fig_results_primary}
\end{figure*}
\begin{figure*}[t!]
    \centering
    \begin{subfigure}[t]{0.25\textwidth}
        \centering
        \includegraphics[width=\linewidth]{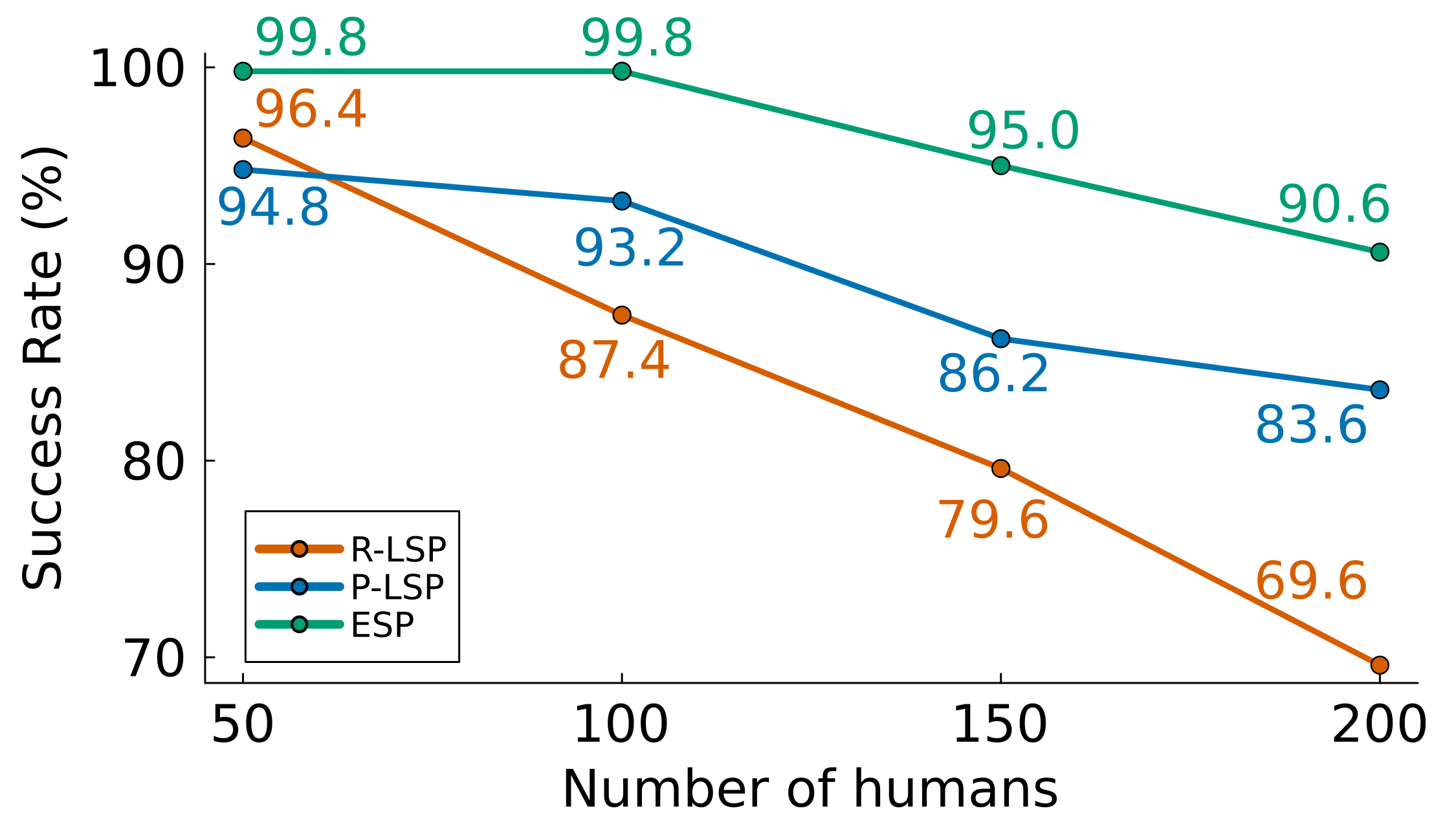}
        \caption{\small Success Rate}
        \label{fig_results_secondary_success_rate}
    \end{subfigure}%
    \hfill
    \begin{subfigure}[t]{0.25\textwidth}
        \centering
        \includegraphics[width=\linewidth]{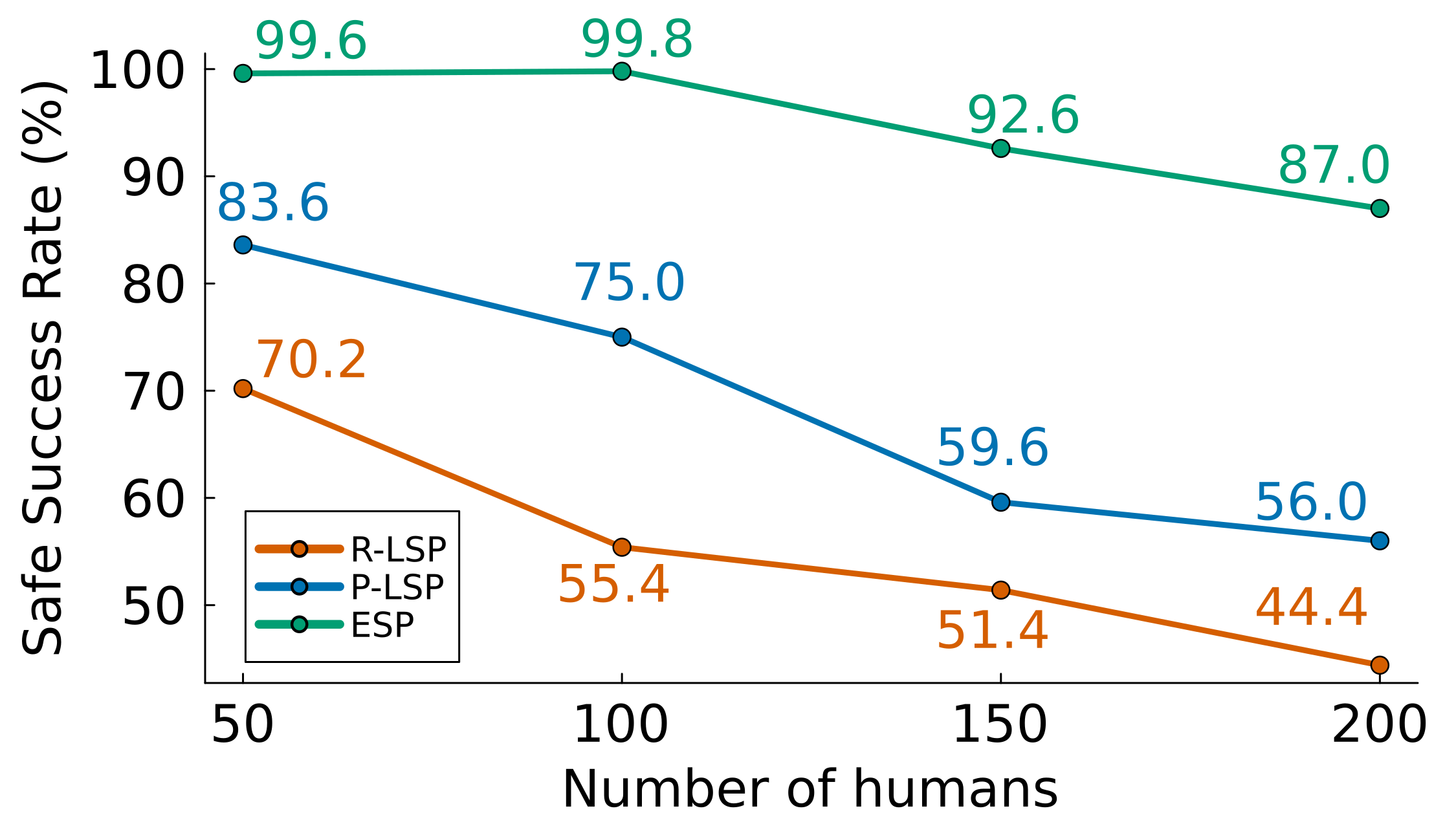}
        \caption{\small Safe Success Rate}
        \label{fig_results_secondary_safe_success_rate}
    \end{subfigure}%
    \hfill
    \begin{subfigure}[t]{0.25\textwidth}
        \centering
        \includegraphics[width=\linewidth]{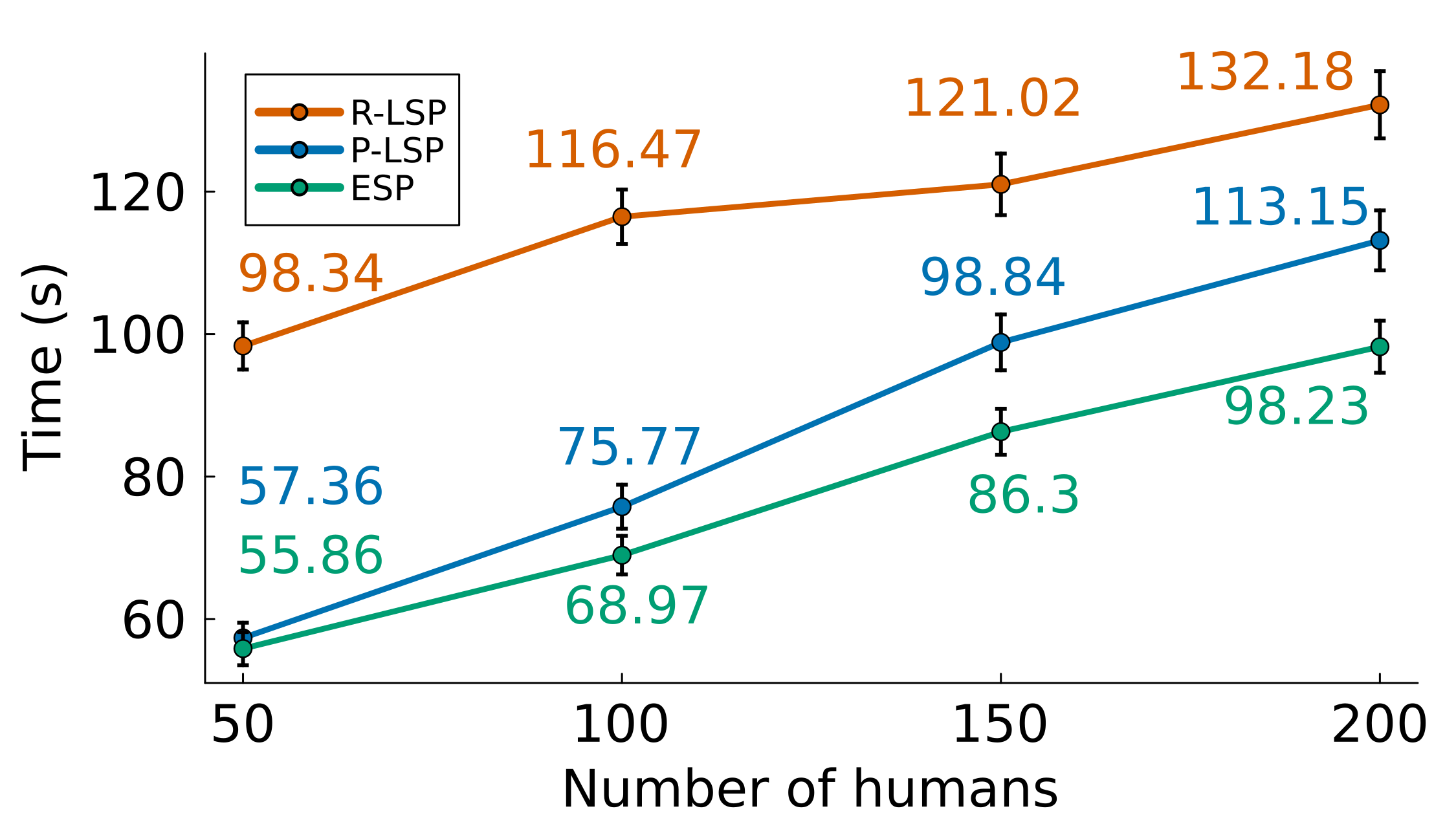}
        \caption{\small Time to Interception}
        \label{fig_results_secondary_interception_time}
    \end{subfigure}
    \caption{\small Secondary study (estimator--planner setting with intermittent target observations).
    Interception success rate, safe-success rate, and time to interception as functions of crowd density when each planner operates using the maximum a posteriori (MAP) target-state estimate.
    }
    \label{fig_results_secondary}
\end{figure*}

\emph{State Propagation:}
During tree search, robot and target dynamics are propagated deterministically (i.e., without process noise), while human motion remains stochastic due to sampled intent hypotheses and bounded motion noise.
Deterministic propagation of robot and target dynamics during search facilitates value propagation under sparse interception rewards and is applied uniformly across all planners.

\emph{Observation Discretization:}
ARDESPOT requires discrete observations, so continuous robot, target and human positions are discretized at a fixed spatial resolution of $2\,\mathrm{m}$. 

\emph{Bounds:} \label{sec_bounds}
ARDESPOT guides its search by maintaining lower and upper bounds at each tree node.
The lower bound is obtained using the rollout policy described in Section ~\ref{sec_rollout_policy}.
The upper bound is the average of per-scenario upper bounds across all scenarios at that node.
For an individual scenario, if the robot is moving and within the safety distance $d_s\!=\!1\mathrm{m}$ of any of the $N$ humans, the upper bound is set to the human collision penalty $R_h$. 
Otherwise, the bound is defined as $\gamma^{t} R_{\mathrm{I}}$, where $t$ is the minimum time required for the robot to intercept the target, assuming it travels at its maximum speed in the absence of humans.
For LSP, $t$ is computed along the reference path. 
For ESP, $t$ is estimated using the same interception procedure used in its rollouts.
This provides an optimistic bound on achievable interception reward under idealized motion without human interference.

\subsection{Planners}
\label{sec_planner_variants}

We compare three planner instantiations that differ in how control authority is exposed to online tree search, while sharing identical belief representations, solver parameters, and reward models.
Each planner operates at 2 Hz, consistent with prior online tree-search approaches for navigation~\cite{bai2015intention,luo2018porca,importance_sampling_despot,gupta_ped_navigation_holonomic}, yielding a planning budget of $0.5\,\mathrm{s}$ per cycle.
For the LSP variants, at most $0.1\,\mathrm{s}$ is allocated to path generation and the remaining time to ARDESPOT tree search. If a new path is not found within $0.1\,\mathrm{s}$, the path from the previous planning cycle is reused. 
For ESP, the full $0.5\,\mathrm{s}$ budget is allocated to tree search.
The three planners are:

\subsubsection{Reactive LSP (R-LSP)}
uses the LSP action space formulation. 
At each planning cycle, Hybrid A$^*$ computes a dynamically feasible path from the robot’s current state to the current target state. 
Human-intent uncertainty is incorporated conservatively during path generation by inflating safety margins.
Hybrid A$^*$ plans using a fixed nominal velocity of $0.5\,\mathrm{m/s}$ and a steering discretization of $19$ equally spaced values in the range $[-\phi_m,\,\phi_m]$.
Details on the path generation process can be found in~\cite{gupta_ped_navigation_holonomic} and~\cite{petereit2012application}.
ARDESPOT then selects the best speed update along this path using the discrete action set
$\delta_v \in \{-0.5,\, 0,\, 0.5\}\,\mathrm{m/s}$.

\subsubsection{Predictive LSP (P-LSP)} extends R-LSP by incorporating target motion prediction into path generation. 
Instead of planning to the current target estimate, Hybrid A$^*$ accounts for target motion by propagating the target state forward under its dynamics during search while simultaneously advancing the robot state. 
This allows the planner to identify a dynamically feasible path that reaches a predicted future target position. 
Tree search remains constrained to the resulting path and selects speed updates as in R-LSP.

\subsubsection{ESP}
uses both steering and speed decisions within the tree search. 
Steering angles are discretized as
$
\{-\phi_m,\,-\tfrac{2\phi_m}{3},\,-\tfrac{\phi_m}{3},\,0,\,\tfrac{\phi_m}{3},\,\tfrac{2\phi_m}{3},\,\phi_m\},
$
with the same set of speed updates $\delta_v \in \{-0.5,\, 0,\, 0.5\}$$\mathrm{m/s}$.
To limit branching during tree search, velocity updates are allowed only when $\phi = 0$ in ESP.
This restriction reduces the effective action space size while preserving representative steering–speed combinations, consistent with common practices in discrete-action online POMDP planning~\cite{gupta_ped_navigation_holonomic}.

\begin{figure*}[t!]
    \centering
    \begin{subfigure}[t]{0.235\textwidth}
        \centering
        \includegraphics[width=\linewidth]{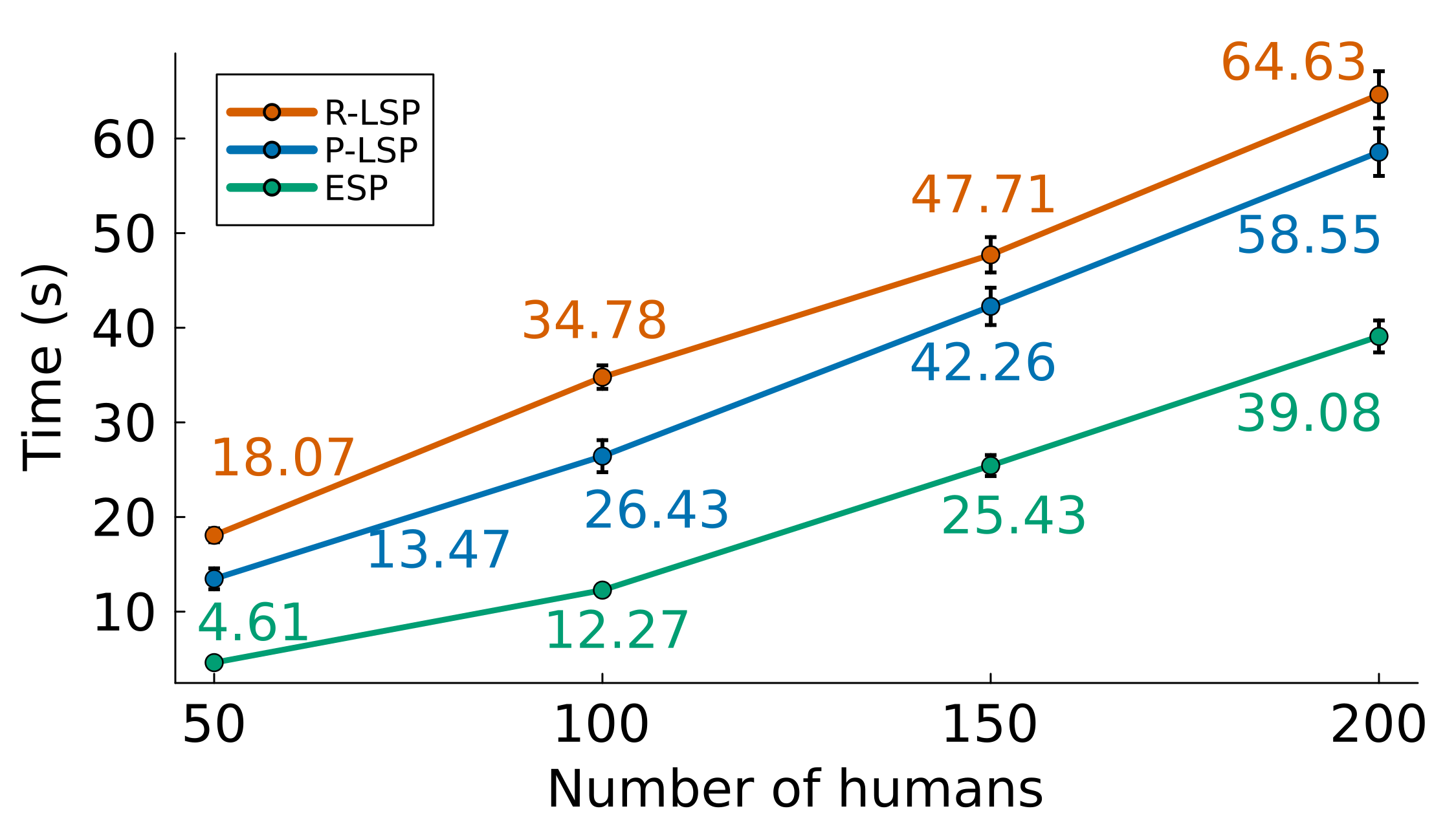}
        \caption{\small Idle Time}
        \label{fig_merged_extra_metrics_idle_time}
    \end{subfigure}%
    \hfill
    \begin{subfigure}[t]{0.235\textwidth}
        \centering
        \includegraphics[width=\linewidth]{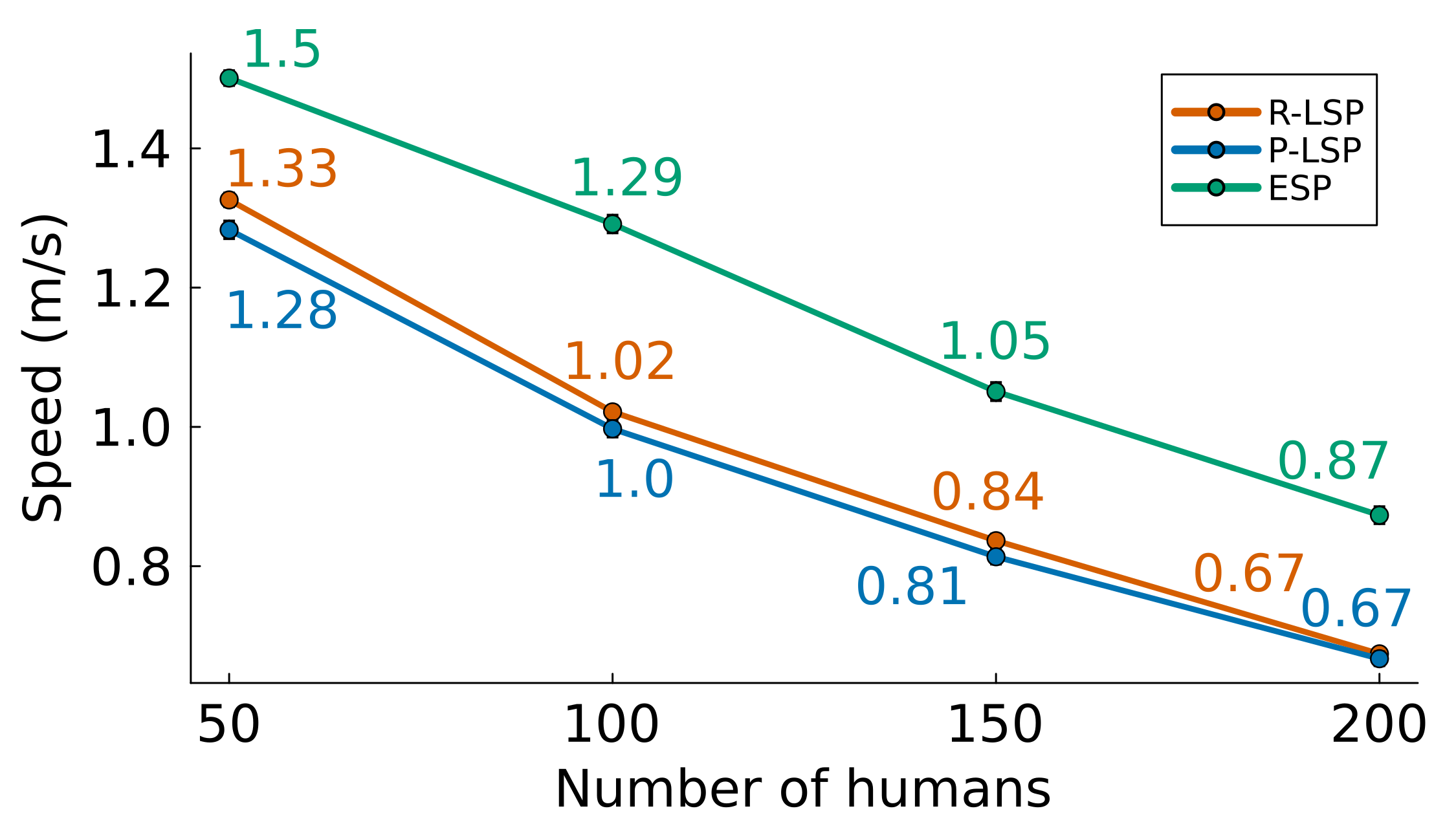}
        \caption{\small Average Speed}
        \label{fig_merged_extra_metrics_average_speed}
    \end{subfigure}%
    \hfill
    \begin{subfigure}[t]{0.235\textwidth}
        \centering
        \includegraphics[width=\linewidth]{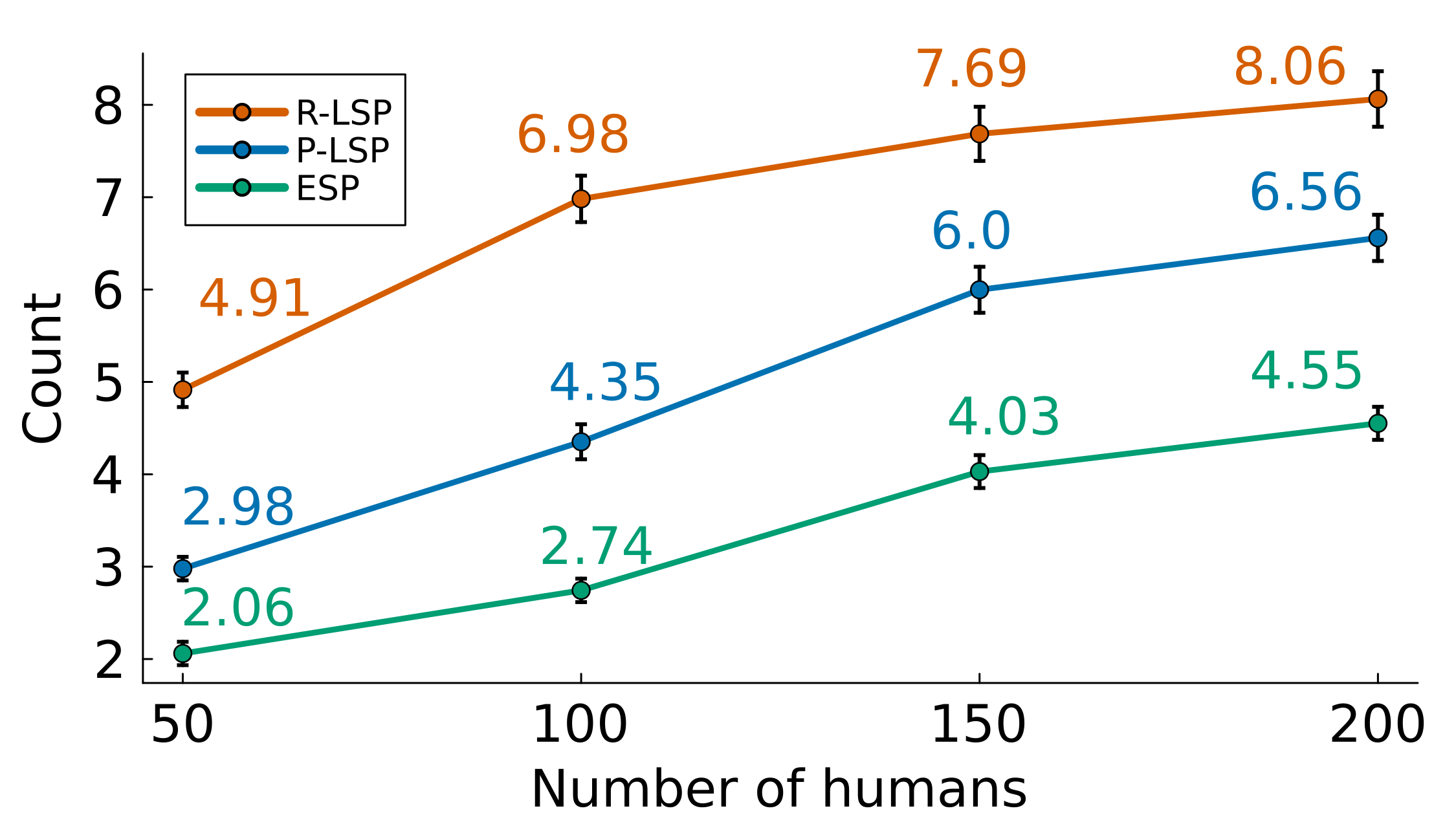}
        \caption{\small Sudden Brake Count}
        \label{fig_merged_extra_metrics_sudden_break}
    \end{subfigure}%
    \hfill
    \begin{subfigure}[t]{0.235\textwidth}
        \centering
        \includegraphics[width=\linewidth]{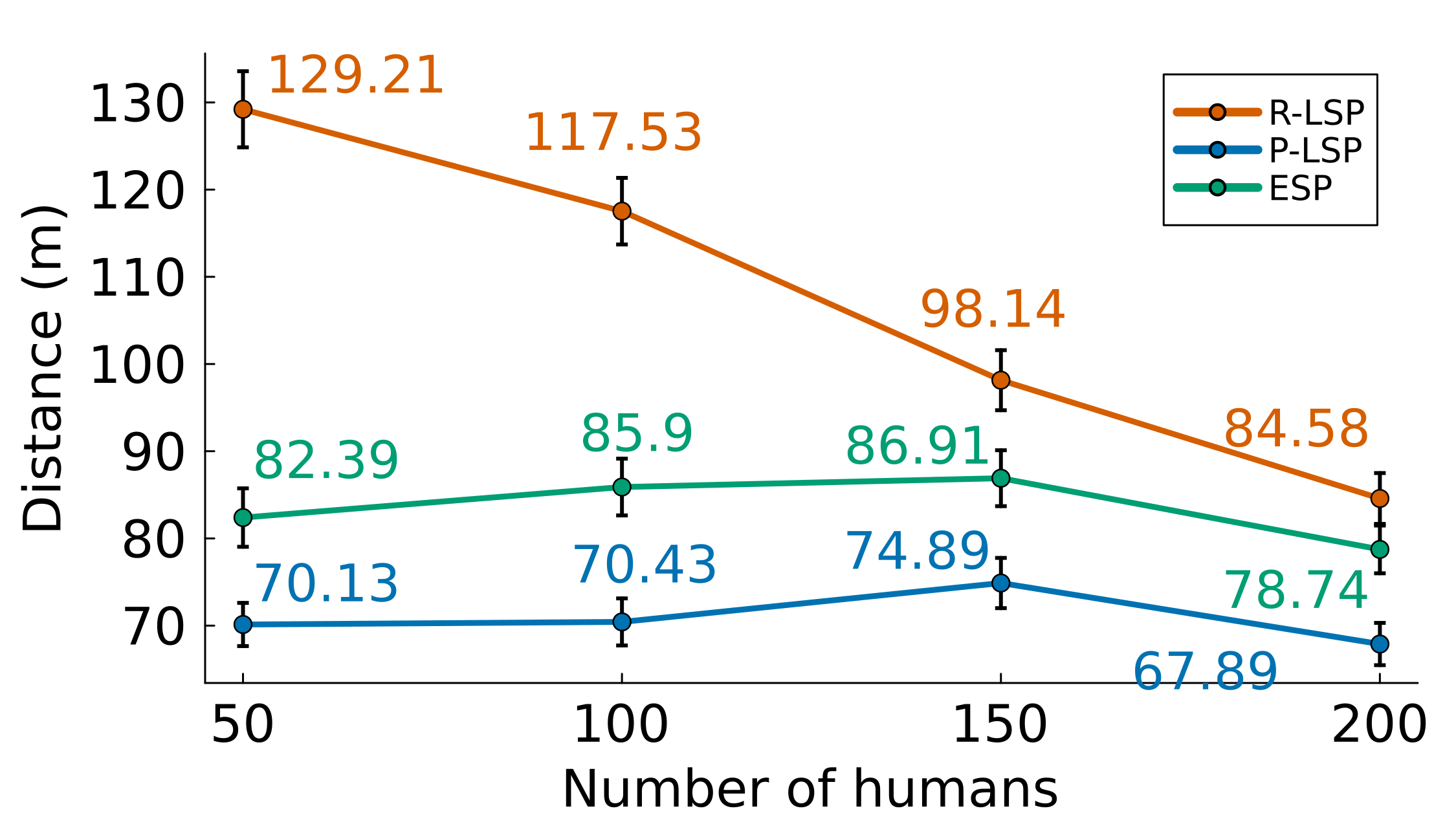}
        \caption{\small Distance Covered}
        \label{fig_merged_extra_metrics_distance_covered}
    \end{subfigure}
    \caption{
    \small
    Motion statistics from the secondary study across increasing crowd densities.
    The figure reports idle time, average speed, sudden braking events, and total distance traveled for ESP, P-LSP, and R-LSP.
}
\label{fig_motion_characteristics}
\end{figure*}
\begin{figure*}[t]
    \centering
    \includegraphics[width=0.99 \linewidth]{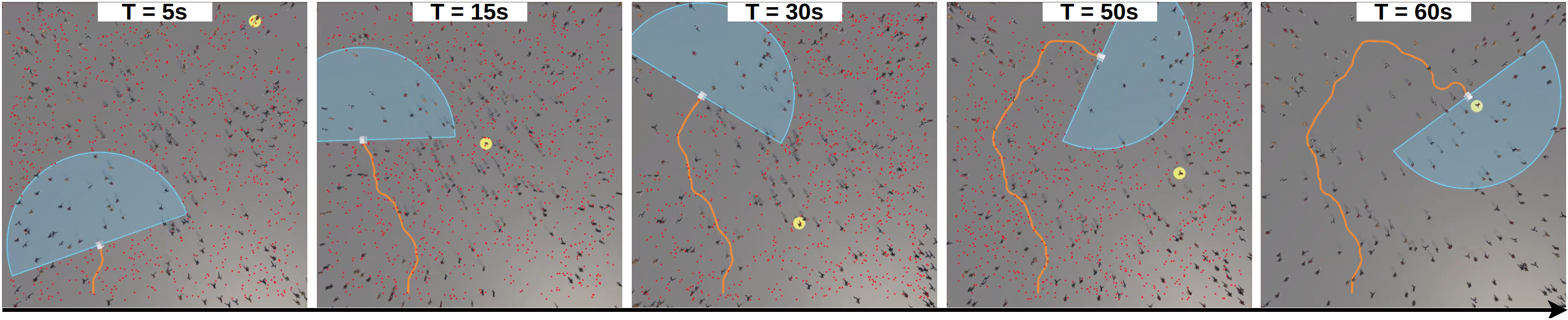}
    \caption{
    \small
    Snapshots from a representative interception trial in the secondary study using ESP at $T=\{5,15,30,50,60\}$\,s.
    The robot (white) navigates through a dense crowd to intercept the target (yellow) while maintaining a belief over the target state.
    Red dots denote particles of the particle-filter belief, black figures represent humans in the environment, and the blue sector illustrates the robot's sensing region.
    The orange curve shows the executed robot trajectory.
    }
    \label{fig_target_tracking_snapshots}
\end{figure*}

\subsection{Experimental Protocol} \label{sec_experiment_protocol}

We conduct $500$ trials for each combination of planner and crowd size ($N_h \in \{50, 100, 150, 200\}$ humans).
Each trial evaluates all planners under identical human and target initializations and motion realizations, enabling paired statistical comparison.
All planners share identical belief representations, solver parameters, and reward models. 
A trial terminates when the robot enters the capture radius $d_c = 1\,\mathrm{m}$ of the target (successful interception), after $300\,\mathrm{s}$ (timeout), or upon collision with the workspace boundary.
A human proximity violation is recorded if the robot, \emph{while moving}, gets within the safety distance $d_s$ of any human; such violations do not terminate the trial but are counted for safety evaluation.
The robot starts near the bottom-left corner of the workspace and the target near the top-right corner to ensure consistent task difficulty and substantial traversal before interception.
Humans are initialized at random positions in the workspace with goals sampled uniformly from the four corners.
To maintain constant crowd density, when a human reaches its goal, it is removed and replaced by a new human spawned along a workspace boundary.
The new human’s goal is sampled from the opposite side to promote motion through the interior.

\emph{Secondary Study:}
The primary study assumes a fully observable target state.
To evaluate robustness to imperfect target information, we consider an estimator--planner architecture in which the target is not directly observed.
A forward-facing detector with $15\,\mathrm{m}$ range and $180^\circ$ field of view provides intermittent measurements of target position $(x_T,y_T)$, with missed detections occurring with probability $p_{\mathrm{m}}=0.1$.
A Sequential Importance Resampling (SIR) particle filter~\cite{particle_filter} with $1000$ particles maintains a belief over the target state.
When a target detection is received, particles are reinitialized around the observed position.
The target heading and speed are estimated from recent consecutive detections.
At each time step, the planner is provided a single target-state hypothesis, represented by the maximum a posteriori (MAP) particle, and treats this hypothesis as the target state for planning.
Scenario initialization is modified accordingly by initializing the target state in all scenarios to this hypothesis; all other components (human belief, solver parameters, action discretization, and rollout policies) remain identical to the main study.
Thus, uncertainty in the target state is handled externally, not modeled within tree search.

%%%%%%%%%%%%%%%%%%%%%%%%%%%%%%%%%%%%%%%%%%%%%%%%%%%%%%%%%%%%%%%%%%%%%%%%%%%%%%%%
\section{Results}

We report interception performance across increasing crowd densities.
A trial is considered \emph{successful} if the robot reaches the target within capture radius $d_c$ before the episode terminates, and \emph{safe-successful} if interception occurs without any human proximity violations.
Success and safe-success rates are computed over all trials.
Motion-related quantitative metrics are 
computed over successful trials only, and error bars denote standard error.
Results are reported for the fully observable target setting (primary study) and the estimator--planner setting (secondary study) described in Sec.~\ref{sec_experiment_protocol}.

\subsection{Interception Performance}

Figs.~\ref{fig_results_primary_success_rate} and \ref{fig_results_secondary_success_rate} report interception success rates for the primary and secondary studies, respectively.
For both studies, the performance of all planners is comparable at the lowest crowd density ($N_h\!=\!50$), but the performance gap widens as density increases.
At $N_h\!=\!200$ in the primary study, ESP and P-LSP have 98\% and 96\% success, while R-LSP drops to 73\%.
Due to estimation uncertainty in the target state, success rate decreases for all planners in the secondary study.
Nevertheless, ESP still maintains around 90\% success while P-LSP and R-LSP drop to 83\% and 69\% at $N_h\!=\!200$.

Fig.~\ref{fig_results_primary_safe_success_rate} reports the safe-success rates for the primary study.
Both LSP variants exhibit significantly lower safe-success than ESP as crowd density increases. 
At $N_h\!=\!200$, ESP retains 97\% safe success, compared to 66\% for P-LSP and 44\% for R-LSP. 
A similar separation is observed in the secondary study (Fig.~\ref{fig_results_secondary_safe_success_rate}).
Incorporating target prediction in path planning improves performance relative to purely reactive path planning, but both LSP variants remain consistently outperformed by ESP.

This separation can be understood through action-space expressiveness during search (Fig.~\ref{fig_planning_trees}).
When steering is not explicitly explored, the planner must rely primarily on speed modulation to navigate around humans, limiting the set of reachable future states at each expansion.
Under dense interactions, this restriction limits the feasible interception maneuvers represented in the search tree.
By expanding both steering and speed together, ESP explores a richer set of trajectories, increasing the probability of identifying collision-free paths to the target as density grows.

\subsection{Motion Characteristics}

To understand how performance differences arise, we examine motion statistics conditioned on successful trials.
Figs.~\ref{fig_results_primary_interception_time} and~\ref{fig_results_secondary_interception_time} show that ESP consistently requires less time to intercept than both LSP variants.
Interception times are higher for all planners in the secondary study than in the primary study due to estimation uncertainty in the target state.
At $N_h\!=\!200$ in the secondary study, ESP requires around $98\,\mathrm{s}$ on average, compared with $114\,\mathrm{s}$ for P-LSP and $132\,\mathrm{s}$ for R-LSP.
This difference is also reflected in motion statistics, with a similar gap in both studies.

Motion statistics from the secondary study are reported in Fig.~\ref{fig_motion_characteristics}.
ESP exhibits lower idle time (Fig.~\ref{fig_merged_extra_metrics_idle_time}), higher average speed (Fig.~\ref{fig_merged_extra_metrics_average_speed}), and fewer sudden braking events (Fig.~\ref{fig_merged_extra_metrics_sudden_break}) than both LSP variants.
These trends indicate that steering-restricted planners exhibit increased hesitation and reactive braking in dense crowds. 
Additionally, Fig.~\ref{fig_merged_extra_metrics_distance_covered} shows the average distance covered by each planner.
ESP travels farther than P-LSP while still achieving shorter interception times.
This suggests that ESP maintains active maneuvering rather than stalling or repeatedly correcting its motion, enabling more consistent progress toward the target.
Representative trajectory snapshots are shown in Fig.~\ref{fig_target_tracking_snapshots}, illustrating continuous progress under ESP.

\emph{Summary:}
These results show that performing online tree search over both heading and speed improves interception reliability, safety, and motion efficiency under dense human interactions.
Restricting robot headings during search limits maneuver diversity and leads to sharp performance degradation as human density increases.
These benefits extend beyond fully observable interception and remain relevant when planning must operate on imperfect target estimates.

%%%%%%%%%%%%%%%%%%%%%%%%%%%%%%%%%%%%%%%%%%%%%%%%%%%%%%%%%%%%%%%%%%%%%%%%%%%%%%%%
\section{Conclusion} \label{section_conclusion}

We presented a POMDP formulation for target interception in crowds, where a robot must intercept a moving target while navigating among intention-driven humans.
Through a controlled comparison between Limited Space Planning (LSP) and Extended Space Planning (ESP) under identical sensing, belief estimation, and computational budgets, we isolated the impact of action-space structure on online tree-search performance.
Our experiments demonstrate that restricting tree search to speed modulation along a precomputed path leads to pronounced performance degradation in safe and reliable interception and motion efficiency as crowd density increases.
In contrast, allowing joint branching over steering and speed within tree search improves interception success, reduces reactive braking, and maintains more consistent forward progress.
These results suggest that action-space design plays a key role in enabling effective online tree-search planning for interception tasks in dense crowds.
Future work will extend the framework to environments with static obstacles, which will require collision-aware, interception-driven rollout policies for ESP.
Efficient parallelized motion-planning methods may enable these rollout policies to be generated within the online planning budget.
%%%%%%%%%%%%%%%%%%%%%%%%%%%%%%%%%%%%%%%%%%%%%%%%%%%%%%%%%%%%%%%%%%%%%%%%%%%%%%%%

\bibliographystyle{IEEEtran}
\bibliography{himanshu_references,kelvin_references,interception_references,zach_references}

@inproceedings{pomcpow,
  title={Online algorithms for POMDPs with continuous state, action, and observation spaces},
  author={Sunberg, Zachary and Kochenderfer, Mykel},
  booktitle={Proceedings of the International Conference on Automated Planning and Scheduling},
  volume={28},
  pages={259--263},
  year={2018}
}

@article{despot,
   title={DESPOT: Online POMDP Planning with Regularization},
   volume={58},
   ISSN={1076-9757},
   DOI={10.1613/jair.5328},
   journal={Journal of Artificial Intelligence Research},
   publisher={AI Access Foundation},
   author={Ye, Nan and Somani, Adhiraj and Hsu, David and Lee, Wee Sun},
   year={2017},
   month=jan, 
    pages={231–266} 
}

@article{importance_sampling_despot,
  title={Importance sampling for online planning under uncertainty},
  author={Luo, Yuanfu and Bai, Haoyu and Hsu, David and Lee, Wee Sun},
  journal={The International Journal of Robotics Research},
  volume={38},
  number={2-3},
  pages={162--181},
  year={2019},
  publisher={SAGE Publications Sage UK: London, England}
}

@inproceedings{vomcpow,
  title={Voronoi progressive widening: Efficient online solvers for continuous state, action, and observation POMDPs},
  author={Lim, Michael H and Tomlin, Claire J and Sunberg, Zachary N},
  booktitle={2021 60th IEEE conference on decision and control (CDC)},
  pages={4493--4500},
  year={2021},
  organization={IEEE}
}

@inproceedings{particle_filter,
  title={Novel approach to nonlinear/non-Gaussian Bayesian state estimation},
  author={Gordon, Neil J and Salmond, David J and Smith, Adrian FM},
  booktitle={IEE proceedings F (radar and signal processing)},
  year={1993},
  organization={IET},
  Ovolume={140},
  Onumber={2},
  Opages={107--113},
}

@inproceedings{gupta_ped_navigation_holonomic,
  title={Intention-Aware Navigation in Crowds with Extended-Space POMDP Planning},
  author={Gupta, Himanshu and Hayes, Bradley and Sunberg, Zachary},
  booktitle={Proceedings of the 21st International Conference on Autonomous Agents and Multiagent Systems},
  pages={562--570},
  year={2022}
}

@inproceedings{gupta2024efficient,
  author    = {Himanshu Gupta},
  title     = {Efficient Continuous Space BeliefMDP Solutions for Navigation and Active Sensing},
  booktitle = {International Conference on Autonomous Agents and Multiagent Systems (AAMAS)},
  OPT_pages     = {2749--2751},
  year      = {2024},
  doi       = {10.5555/3635637.3663275}
}

@inproceedings{bai2015intention,
  title={Intention-aware online POMDP planning for autonomous driving in a crowd},
  author={Bai, Haoyu and Cai, Shaojun and Ye, Nan and Hsu, David and Lee, Wee Sun},
  booktitle={2015 IEEE International Conference on Robotics and Automation},
  OPT_pages={454--460},
  year={2015},
  organization={IEEE}
}

@article{luo2018porca,
  title={Porca: Modeling and planning for autonomous driving among many pedestrians},
  author={Luo, Yuanfu and Cai, Panpan and Bera, Aniket and Hsu, David and Lee, Wee Sun and Manocha, Dinesh},
  journal={IEEE Robotics and Automation Letters},
  Ovolume={3},
  Onumber={4},
  Opages={3418--3425},
  year={2018},
  publisher={IEEE}
}

@inproceedings{lee2020magic,
  title     = {{MAGIC}: Learning Macro-Actions for Online {POMDP} Planning},
  author    = {Lee, Yiyuan and Cai, Panpan and Hsu, David},
  booktitle = {Proceedings of Robotics: Science and Systems},
  year      = {2021},
  OPT_address   = {Virtual},
  OPT_month     = jul,
  doi       = {10.15607/RSS.2021.XVII.041}
}

@article{jin2025hi,
  title={Hi-Drive: Hierarchical POMDP Planning for Safe Autonomous Driving in Diverse Urban Environments},
  author={Jin, Xuanjin and Zeng, Chendong and Zhu, Shengfa and Liu, Chunxiao and Cai, Panpan},
  journal={IEEE Robotics and Automation Letters},
  year={2025},
  publisher={IEEE}
}

@inproceedings{petereit2012application,
  title={Application of hybrid A* to an autonomous mobile robot for path planning in unstructured outdoor environments},
  author={Petereit, Janko and Emter, Thomas and Frey, Christian W and Kopfstedt, Thomas and Beutel, Andreas},
  booktitle={ROBOTIK 2012; 7th German conference on Robotics},
  pages={1--6},
  year={2012},
  organization={VDE}
}

@article{egorov2017pomdps,
  author  = {Maxim Egorov and Zachary N. Sunberg and Edward Balaban and Tim A. Wheeler and Jayesh K. Gupta and Mykel J. Kochenderfer},
  title   = {{POMDP}s.jl: A Framework for Sequential Decision Making under Uncertainty},
  journal = {Journal of Machine Learning Research},
  year    = {2017},
  volume  = {18},
  number  = {26},
  pages   = {1-5}
}

@inproceedings{letsdrive,
  title     = {{LeTS-Drive}: Driving in a Crowd by Learning from Tree Search},
  author    = {Cai, Panpan and Luo, Yuanfu and Saxena, Aseem and Hsu, David and Lee, Wee Sun},
  booktitle = {Proceedings of Robotics: Science and Systems},
  year      = {2019},
  Oaddress   = {Freiburg im Breisgau, Germany},
  Omonth     = jun,
  doi       = {10.15607/RSS.2019.XV.018}
}

@article{pomdp_complexity_papadimitriou_1987,
  title={The Complexity of {M}arkov Decision Processes},
  author={Papadimitriou, Christos H. and Tsitsiklis, John N.},
  journal={Mathematics of Operations Research},
  OPT_volume={12},
  OPT_number={3},
  OPT_pages={441--450},
  year={1987},
  publisher={{INFORMS}}
}

@article{swiechowski2023monte,
  title={Monte Carlo tree search: A review of recent modifications and applications},
  author={{\'S}wiechowski, Maciej and Godlewski, Konrad and Sawicki, Bartosz and Ma{\'n}dziuk, Jacek},
  journal={Artificial Intelligence Review},
  volume={56},
  number={3},
  pages={2497--2562},
  year={2023},
  publisher={Springer}
}

@article{hu2022sharp,
  title={Sharp: Shielding-aware robust planning for safe and efficient human-robot interaction},
  author={Hu, Haimin and Nakamura, Kensuke and Fisac, Jaime F},
  journal={IEEE Robotics and Automation Letters},
  volume={7},
  number={2},
  pages={5591--5598},
  year={2022},
  publisher={IEEE}
}

@article{mustafa2024racp,
  title={RACP: Risk-aware contingency planning with multi-modal predictions},
  author={Mustafa, Khaled A and Ornia, Daniel Jarne and Kober, Jens and Alonso-Mora, Javier},
  journal={IEEE Transactions on Intelligent Vehicles},
  year={2024},
  publisher={IEEE}
}

@article{peters2024contingency,
  title={Contingency games for multi-agent interaction},
  author={Peters, Lasse and Bajcsy, Andrea and Chiu, Chih-Yuan and Fridovich-Keil, David and Laine, Forrest and Ferranti, Laura and Alonso-Mora, Javier},
  journal={IEEE Robotics and Automation Letters},
  OPT_volume={9},
  OPT_number={3},
  OPT_pages={2208--2215},
  year={2024},
  publisher={IEEE}
}

@inproceedings{xu2023benchmarking,
  title={Benchmarking reinforcement learning techniques for autonomous navigation},
  author={Xu, Zifan and Liu, Bo and Xiao, Xuesu and Nair, Anirudh and Stone, Peter},
  booktitle={IEEE International Conference on Robotics and Automation},
  OPT_pages={9224--9230},
  year={2023},
  OPT_organization={IEEE}
}

@inproceedings{raj2024rethinking,
  title={Rethinking social robot navigation: Leveraging the best of two worlds},
  author={Raj, Amir Hossain and Hu, Zichao and Karnan, Haresh and Chandra, Rohan and Payandeh, Amirreza and Mao, Luisa and Stone, Peter and Biswas, Joydeep and Xiao, Xuesu},
  booktitle={2024 IEEE International Conference on Robotics and Automation (ICRA). IEEE},
  year={2024}
}

@inproceedings{vaskov2019not,
  title={Not-at-fault driving in traffic: A reachability-based approach},
  author={Vaskov, Sean and Larson, Hannah and Kousik, Shreyas and Johnson-Roberson, Matthew and Vasudevan, Ram},
  booktitle={IEEE Intelligent Transportation Systems Conference},
  OPT_pages={2785--2790},
  year={2019},
  OPT_organization={IEEE}
}

@article{lauri2022partially,
  title={Partially observable markov decision processes in robotics: A survey},
  author={Lauri, Mikko and Hsu, David and Pajarinen, Joni},
  journal={IEEE Transactions on Robotics},
  volume={39},
  number={1},
  pages={21--40},
  year={2022},
  publisher={IEEE}
}

@article{helbing1995social,
  title={Social force model for pedestrian dynamics},
  author={Helbing, Dirk and Molnar, Peter},
  journal={Physical review E},
  volume={51},
  number={5},
  pages={4282},
  year={1995},
  publisher={APS}
}

@article{pliska2024towards,
  title={Towards safe mid-air drone interception: Strategies for tracking \& capture},
  author={Pliska, Michal and Vrba, Matou{\v{s}} and B{\'a}{\v{c}}a, Tom{\'a}{\v{s}} and Saska, Martin},
  journal={IEEE robotics and automation letters},
  volume={9},
  number={10},
  pages={8810--8817},
  year={2024},
  publisher={IEEE}
}

@INPROCEEDINGS{Qu2022,
  author={Qu, Chendi and He, Jianping and Li, Jialun and Fang, Chongrong and Mo, Yilin},
  booktitle={2022 American Control Conference (ACC)}, 
  title={Moving Target Interception Considering Dynamic Environment}, 
  year={2022},
  volume={},
  number={},
  pages={1194-1199},
  doi={10.23919/ACC53348.2022.9867177}}

@article{zheng2021time,
  title={Time-optimal Guidance for Intercepting Moving Targets by {Dubins} Vehicles},
  author={Zheng, Yuan and Chen, Zheng and Shao, Xueming and Zhao, Wenjie},
  journal={Automatica},
  OPT_volume={128},
  OPT_pages={109557},
  year={2021},
  publisher={Elsevier}
}

@article{gupta2022shortest,
  title={Shortest Dubins paths to intercept a target moving on a circle},
  author={Gupta Manyam, Satyanarayana and Casbeer, David W and Von Moll, Alexander and Fuchs, Zachariah},
  journal={Journal of Guidance, Control, and Dynamics},
  Ovolume={45},
  Onumber={11},
  Opages={2107--2120},
  year={2022},
  publisher={American Institute of Aeronautics and Astronautics}
}

@inproceedings{schlotfeldt2019maximum,
  title={Maximum information bounds for planning active sensing trajectories},
  author={Schlotfeldt, Brent and Atanasov, Nikolay and Pappas, George J},
  booktitle={IEEE/RSJ International Conference on Intelligent Robots and Systems},
  OPT_pages={4913--4920},
  year={2019},
  organization={IEEE}
}

@inproceedings{yang2023policy,
  title={Policy Learning for Active Target Tracking over Continuous $ SE (3) $ Trajectories},
  author={Yang, Pengzhi and Koga, Shumon and Asgharivaskasi, Arash and Atanasov, Nikolay},
  booktitle={Learning for Dynamics and Control Conference},
  OPT_pages={64--75},
  year={2023},
  organization={PMLR}
}

@inproceedings{olsen2021visibility,
  title={A visibility roadmap sampling approach for a multi-robot visibility-based pursuit-evasion problem},
  author={Olsen, Trevor and Tumlin, Anne M and Stiffler, Nicholas M and O’Kane, Jason M},
  booktitle={2021 IEEE International Conference on Robotics and Automation (ICRA)},
  pages={7957--7964},
  year={2021},
  organization={IEEE}
}

@inproceedings{olsen2022robust,
  title={Robust-by-design plans for multi-robot pursuit-evasion},
  author={Olsen, Trevor and Stiffler, Nicholas M and O'Kane, Jason M},
  booktitle={2022 International Conference on Robotics and Automation (ICRA)},
  pages={10716--10722},
  year={2022},
  organization={IEEE}
}

@article{bertsekas2005dynamic, title={Dynamic Programming and Suboptimal Control: A Survey from ADP to MPC*},
OPT_volume={11}, 
ISSN={0947-3580},
DOI={10.3166/ejc.11.310-334}, 
OPT_number={4},
journal={European Journal of Control},
author={Bertsekas, Dimitri P.}, year={2005}, 
OPT_month=jan,
OPT_pages={310–334} }

\end{document}